\documentclass[conference]{IEEEtran}
\IEEEoverridecommandlockouts
\usepackage{cite}
\usepackage{romannum}
\usepackage{amsmath,amssymb,amsfonts}
\usepackage{algorithmic}
\usepackage{epstopdf}
\usepackage{graphicx}
\usepackage{textcomp}
\usepackage{xcolor}
\def\BibTeX{{\rm B\kern-.05em{\sc i\kern-.025em b}\kern-.08em
    T\kern-.1667em\lower.7ex\hbox{E}\kern-.125emX}}
\begin{document}

\title{Active learning with binary models for real time data labelling \\}

\author{\IEEEauthorblockN{Ankush Deshmukh}
\IEEEauthorblockA{\textit{Electronics and Communication} \\
\textit{NIT Surathkal}\\
Manglore, India \\
ankushsanjaydeshmukh.202sp003@nitk.edu.in}
\and
\IEEEauthorblockN{Bhargava B C}
\IEEEauthorblockA{\textit{Electronics and Communication} \\
\textit{NIT Surathkal}\\
Manglore, India \\
bhargavabc.203sp001@nitk.edu.in}

\and

\IEEEauthorblockN{{A V Narasimhadhan}}
\IEEEauthorblockA{\textit{Electronics and Communication} \\
\textit{NIT Surathkal}\\
Manglore, India \\
dhan257@gmail.com}}
\maketitle

\begin{abstract}
Machine learning (ML) and Deep Learning (DL) tasks primarily depend on data. Most of the ML and DL applications involve supervised learning which requires labelled data. In the initial phases of ML realm lack of data used to be a problem, now we are in a new era of big data. The supervised ML algorithms require data to be labelled and of good quality. labelling tasks may require a large amount of money and time investment. Data labelling may require a skilled person who may charge high for this task, consider the case of the medical field or the data is in bulk that may require a lot of people assigned to label it. Amount of data that is well enough for training needs to be known, money and time can not be wasted to label the whole data. This paper mainly aims to propose a strategy that helps in labelling the data along with oracle in real-time. With balancing "on" model contribution for labelling is 89 and 81.1 for furniture type and intel scene image data sets respectively. Further with balancing being kept "off" model contribution is found to be 83.47 and 78.71 for furniture type and flower data sets respectively.

\end{abstract}

\section{Introduction}
In the field of Convolutional neural networks (CNN), we have seen many new kinds of research and advancements related to model architectures and mathematical explanations behind them. It is mostly been statistical and less methodological. Our main aim is to use existing architectures and achieve efficiency in labelling tasks. A problem that we want to solve is labelling unlabelled data and train models with as least number of data samples as possible to save the training time.\\

Attempts have been made to build such a methodology to reduce the efforts required for training ML, DL models with the least number of data samples, which directly means a reduction in time and cost required for the task. Three main pillars in our attempt are pre-existing well trained models, selective sampling based active learning algorithms, and most important of all individualized binary models for classification (one verse all). 
\subsection{Pretrained models}
Convolutional neural networks can be broken down into two stages, the convolutional network stage with filters which learns to recognize the shapes in the image, and the classifier stage which is responsible for classification tasks based on the features extracted by the convolutional stage. The first stage of the network requires adjusting its filter weights to learn feature extraction and to tune those filter weights need data. The first hurdle is solved by using the pre-trained feature extractor layers which are open source available on the Tensorflow hub. A model can be easily selected which is relevant to the task. This is also known as transfer learning. Taking the pretrained model's weights saves a lot of data samples required for training.\\

\subsection{Active learning} 
Active learning is a subset of machine learning in which the model suggests the most relevant data points from the dataset. The aim is in place of passive model learning there can be
model involvement to select these samples from the dataset. There has been a lot of statistical research in this field. There are different ways to get these desired samples with least confident sampling, marginal sampling, entropy sampling. This is a significant part of methodology because we do not want random samples to be thrown in model training, important
samples selection can be done using favorable strategy. The model’s inference suggests us the relevant samples it requires.
\subsection{One versus all}
A single model can be used to do a multi-class classification task, although the real issue a single model faces is class imbalance. When there is skewness in data there comes the issue of overfitting, in this case, it is more reliable to go for individualized binary models. The steps are simple, every single class will have a binary model associated with it, which learns to classify objects that belong to a particular class versus the other classes. The binary models seem to be more helpful in the application we want to build rather than a single model.
\section{Related works}
Related works can be split into two parts. The first one is work-related to active learning and the second one is work-related to binary classification. Let us have a look at active learning-based works.

Pioneer significant work started with the proposal of a pool-based active learning model having an incremental decision tree as the basic structure \cite{b1}. The proposed method in \cite{b1} helped in the removal of ambiguity in the unsampled pool. The best accuracy is found out to be 92.50. Performance of active learning the framework gets affected due to an imbalance in class distribution. A detailed discussion on factors affecting active learning and the reason for imbalanced data affecting the performance of active learning was made in \cite{b2}. The hierarchical clustering method is used. The basic classifier used is a weighted extreme learning model and further proposed architecture is deduced out of this and named as Active online weighted extreme learning model (AOW-ELM). Detailed running time and performance comparisons were made for 32 binary class data sets with varieties of imbalance ratios and the proposed AOW-ELM was found to be better than many of pre-existing state-of-the-art model for active learning \cite{b2}.\\

Arabic NADA news data was labelled using active learning and an accuracy of 99 percent was obtained when just 17.8 percentage of samples were trained, active learning was proposed with Jaccard similarity classifier \cite{b3}. Active learning found its way to the classification of visual imperfections in the industry through \cite{b4}. In reference \cite{b4} three different methods were proposed, first one is auto labelling enhanced active learning (ALEAL) which was found to be cost-efficient and was working based on deep convolution neural networks, followed by diverse cost-efficient query strategies (DCEQS), followed with attention-based similarity measurement network (ASMN), ALEAL was used to reduce the human intervention during labelling, DCEQS helps in the selection of samples with high performance and further ASMN is used to
determine the similarity between labelled and unlabelled samples. A deep supervised U-Net-based active learning framework was proposed for weak and strong labelers in biomedical image segmentation, further models performance was evaluated using the Internet of medical things (IMOT) \cite{b5}.

Binary classification has got its significance in a wide range of areas. Text classification method with binary classifier enabled with difference frequency method was proposed and F1 score was found to be improving from 88.12 percent to 91.96 \cite{b6}. Performance of binary tree-based support vector machine (SVM) architecture on simple training data sets were evaluated and it was found that binary classifier initially separates the most efficiently distributed class from the whole distribution and thus performs better. SVM based on a binary tree was seen performing better due to its ability to construct tree according to the distribution of data set \cite{b7}. Cancer images with metastatic conditions were classified using binary classifiers with Resnet architecture, further results from the model were found to perform better than that of vgg16 and vgg19 based networks \cite{b8}. The main thing was scan detection of digital pathology being converted into a problem of binary classification. From the above-mentioned papers, one thing can be concluded, binary classifiers gives extensive performance over a wide range of areas, and their applications.\\

One versus all (OVA) problem in multi-class classification is always seen to be performing better. The problem of best suitable kernel selection for one versus all classification was taken and solved with the help of a set of error bounds which was minimized for various conditions \cite{b9}. These kernels were then imposed on top of specialized SVM classifiers. Further, the multi-class classification problem was solved with a decision tree-based one versus all classifiers \cite{b10}. An OVA-based decision tree was trained on 20 data streams and was seen performing better than preexisting models. Weight updating and training were found to be faster and the accuracy of classification was also higher. Models performance was theoretically and empirically proved. Merits of OVA classifiers such as high accuracy, getting adopted to new class labels found in between the class were discussed in detail \cite{b10}. Demerits of OVA are, not being able to handle concept change, classification accuracy gets affected by the data set having an imbalanced class label. Hence in this paper, we try to solve these problems by proposing a most efficient algorithm based on active learning with pre-trained models and buffer.
\section{Datasets and preprocessing}
The datasets considered in this experimental setup are flowers dataset, car damage or not damage dataset, mask-No-mask dataset, yoga-pose dataset, car-logos dataset, furniture type dataset, intel seg dataset. The images were reshaped as 244x244 and normalized between 0 and 1, before sending to the feature extractor layer.
\section{Methodological reasoning}
Fig. 1 shows the flow diagram for the proposed method. Various blocks included in the figure need to be elaborate with reasoning, like the use of one versus all, selective sampling method, buffer strategy in the algorithm. The reasons are as follows,\\
\begin{itemize}
    \item A single large model with an softmax layer at end with neurons equal to classes is initialized, new classes can not be included easily in between the labelling task. The model will have to be built again with new weights and we will lose all the training. To avoid this binary models are used. Even when the user demands a new class in between, a new binary model can be easily built for particular class without affecting the old classes and their parameters. 
    \item User can not always be constrained to label classes on equal counts. which is necessary to avoid overfitting in the single model scenario. Whereas binary models are proven to be more resistant towards imbalance.
    \item Active learning helps in the selection of important data for the model, further it assumes the data to be balanced and equally distributed across the classes. Different approach is used to select the important samples. In this strategy, buffer is being added. Buffer accumulates incorrect predictions by the model and their correct labels by the user. Once the buffer is full, samples are sorted in increasing order of their incorrect confidence across the classes. Idea is to pick samples that strategy predicted with higher confidence and are incorrect. Then train the models with these samples, this is penalizing with most incorrect attempts by the model. In the experiment top 10 samples from each available class in the buffer are picked.
    \item A single difficult class may affect all others in single model architecture. In binary mode, all models are independent of each other. Hence cost functions of each classes are independent and do not affect across the classes.
    \item Since the binary models are independent, while training these models can be trained in parallel. In this experiment it is sequential.
\end{itemize}
\subsection{Methodology flow}
\begin{figure}[htbp]
\centerline{\includegraphics[scale=0.35]{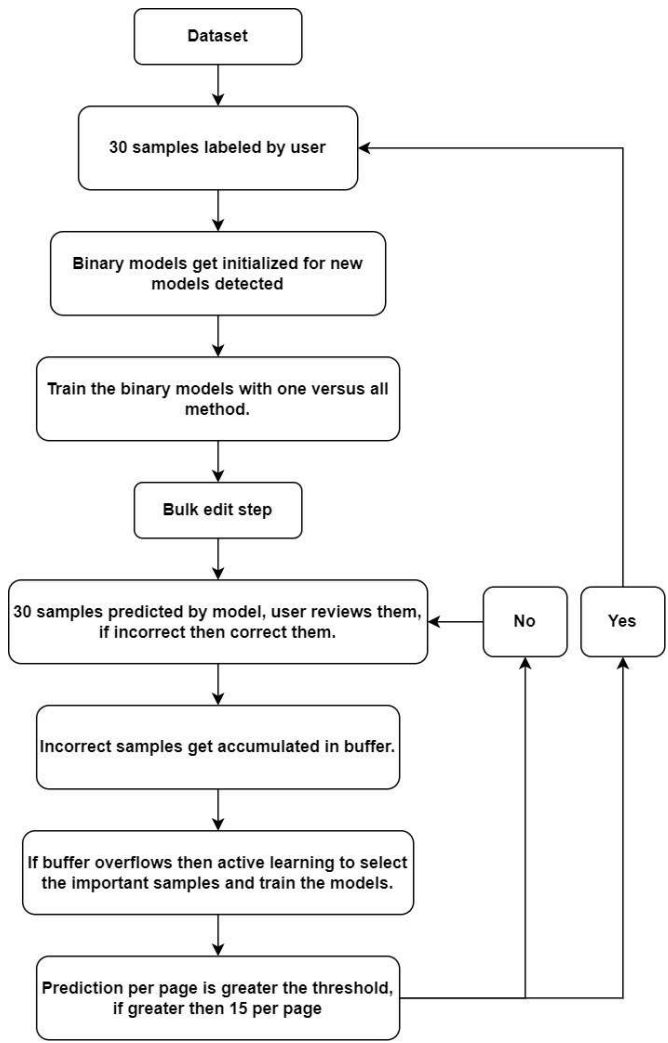}}
\caption{Flow diagram for proposed algorithm}
\label{fig .1}
\end{figure}
\begin{itemize}
    \item Random 30 images are labelled by the user.
    \item Each unique label builds a new binary model for that specific class as and when detected. The model is a simple TensorFlow hub feature extractor layer followed by a dense layer of 50 neurons and ends with a single neuron with sigmoid activation, optimizer used is sgd. The underlying feature extractor used here is MobileNet-V3-large.
    \item The data is sent for training each model. Before sending the data for training for a particular model, data is processed to change the labels as positive or negative samples.
    \item Models are trained with standard configuration, batch size of 10 with epochs of 20.
    \item Bulk edits stage - next 30 sample's labels are predicted by models based on a model whose inference confidence is maximum.
    \item User can correct the wrong predictions by model and accept all remaining. These incorrect predictions are stored in a buffer along with the correct labels by the user.
    \item Constant check is made for mistake greater than 15 per iteration among 30 samples or the buffer size is full. Buffer size taken here is the number of classes x 20.  
    \item  In case buffer size is full, use active selection to select important samples for training the models from the buffer. The samples from each class are separated and among each class, we select the top 10 samples.
    \item In case a batch makes greater than 15 mistakes, go to the initial stage of labelling random 30 samples.
    \item Else continue bulk edit stage of next 30 samples labelled by models.
\end{itemize}
Balancing strategy is an experimental parameter that allows having data augmentation before data goes into each model's training, balancing positive and negative samples which is the most probable scenario to happen in binary models strategy. Augmentation techniques included herein general acts like flipping horizontally or vertically, stretching, etc. Sorting is another experimental parameter used in a buffer while training them according to their confidence which can be high to low or low to high. 
\section{Results}
Table \Romannum{1} and Table \Romannum{2} show simulation results related to some general datasets taken from the kaggle website. The table shows the size of data considered after cleaning, the number of classes in that dataset. Models contribution is the portion of data correctly labelled by models. Time taken is an important parameter since it needs to be realtime, it is represented in minutes, training time is a subset of time taken for the full script. Time taken for script also includes time taken for inference before labelling by models. Buffer sorting is high to low. Table \Romannum{1} shows the results when balancing is kept off and Table \Romannum{2} shows the result with balancing kept on. Results show improvement in the outcomes when balancing strategy is used and it also doesn't affect the overall time taken as such compared to when balancing is kept off.\\

Given figures are the results for outcomes for classification of images into damaged car or not damaged car with balancing strategy on. Fig. 2. shows model contribution versus the number of iterations for . Across the iterations, the user and model are to label a batch of 30 samples together. The value at a particular iteration tells you the number of samples correctly labelled by the models. Fig. 3. shows the time taken for training versus the training number. Throughout the task, there are two possibilities of training to happen, either the model predictions are incorrect for more than half of a particular batch size or the buffer is full. These time values can be high because of a larger count of classes resulting in a higher count of individual models. The training in experiments happens sequentially, this is the reason behind some simulations taking more time in training. Due to the independence of the models, with a special setup, these models can also be trained parallel and time can be saved. Fig. 4 is more intuitive way to understand the effectiveness of the method, blue is the amount correctly labelled by the models and orange is the labelling done by the user.
\begin{table}[ht]
\caption{Comparison with Balancing off}
\renewcommand{\arraystretch}{1.35}
\setlength{\tabcolsep}{5pt} 
\scriptsize 
\begin{tabular}{ |p{1.8cm}|p{.7cm}|p{0.55cm}|p{1cm}|p{1cm}|p{1.2cm}|}
 \hline
\textbf{Dataset} & \textbf{No of. samples}&\textbf{classes}  & \textbf{Model contribution} & \textbf{Time taken for training(mins)} & \textbf{Time taken for full script(mins)} \\
\hline

 Flower &2253 &5 & 78.71 & 1.8 &13\\
 
 \hline
 Car damage & 1655 & 2 & 76 & 0.6 &4\\
  \hline
  Furniture type &3625 & 5 & 83.47 & 1.5&20\\
  \hline
  Mask-No mask &892  & 2 & 45 & 0.6&2.1\\
  \hline
  Yoga pose &790 &5  & 60.25 & 1.6&5.4\\
  \hline
Car logos &2024  & 8 & 55 &5.1 &20\\
  \hline
 Intel scene image &8159 & 6 & 75.85 & 4.6&57\\
 \hline

\end{tabular}
\end{table}

\begin{table}[ht]
\caption{Comparison with balancing on}
\renewcommand{\arraystretch}{1.35}
\setlength{\tabcolsep}{5pt} 
\scriptsize 
\begin{tabular}{ |p{1.8cm}|p{.7cm}|p{0.55cm}|p{1cm}|p{1cm}|p{1.2cm}|}
 \hline
\textbf{Dataset} & \textbf{No of. samples}& \textbf{classes}  & \textbf{Model contribution} & \textbf{Time taken for training(mins)} & \textbf{Time taken for full script(mins)}\\
\hline

Flower & 2253 &5 & 79.2 & 1.3 & 9.9\\
 \hline
 Car damage &  1655 & 2 &  75.81 &  0.48 & 3\\
  \hline
  Furniture type & 3625 & 5 &  89 & 1.2& 15\\
  \hline
 Mask-No mask & 892  & 2 &  50 &  0.55& 1.7\\
  \hline
 Yoga pose & 790 & 5  & 64.8 & 1& 3.7\\
  \hline
 Car logos & 2024  & 8 & 54.52 & 5.14 & 16\\
  \hline
 Intel scene image & 8159 & 6 & 81.1 & 5& 42.3\\
 \hline

\end{tabular}
\end{table}
\begin{figure}[htbp]
\centerline{\includegraphics[scale=0.5]{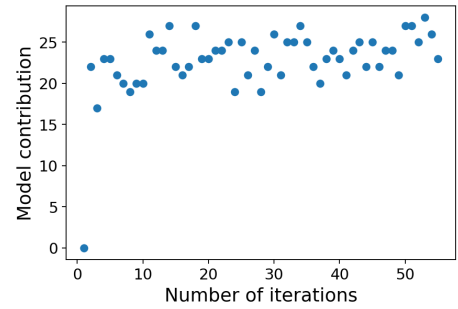}}
\caption{Model contribution in each iteration}
\label{fig .2}
\end{figure}
\begin{figure}[htbp]
\centerline{\includegraphics[scale=0.5]{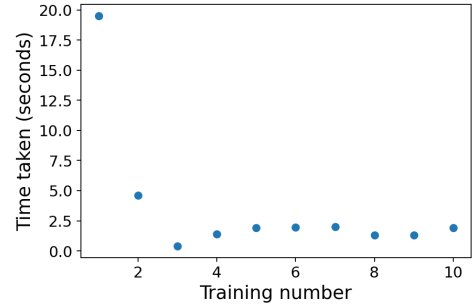}}
\caption{Training time taken across the count of training counts.}
\label{fig .3}
\end{figure}
\begin{figure}[htbp]
\centerline{\includegraphics[scale=0.344]{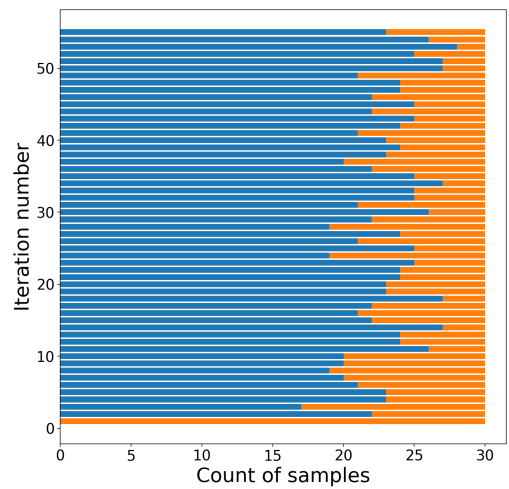}}
\caption{Model vs User labelling.}
\label{fig .4}
\end{figure}

\section{Conclusion}
The strategy is tested on datasets from kaggle with labels provided for running simulation script, to understand the way it helps across segments. The method appears to be helping in most of the scenarios, even in the worst-case model accounted for 50 percent of help which is the same as a person doing double the capability. The number of times the model is trained and the overall time taken is the major thing to look at, considering the help model gives in real time.\\ 
Major advantages of the proposed method are,
\begin{itemize}
\item These binary models generated across labelling tasks can be saved and used on the future dataset or can act as a baseline for the search of a new model on which users want to train using this labelled data.
\item User can easily add extra class in between of labelling task without disturbing previous models learning since the models are independent they can easily be generated.
\item Data skewness shows less effect on individualized binary models and hence the user is not a constraint in labelling the data in equal amounts for each class, to cover up an imbalance in binary models balancing strategies can also be added as per need.
\item Selection strategy among all samples in buffer helps in saving the time taken for training models. Selective sampling helps in including only important samples that can lead to better learning and with reduced training time rather than sending all the data for training.
\end{itemize}
The base feature extractor used in this experiment is MobileNet-V3-large, the outcome of the experiment was majorly based on these features extracted. Hence as the future scope of this experimentation, the efforts will be to find a way to rank among various feature vectors in real time while the user labels to select the best-suited model that can yield a higher contribution.

\end{document}